\documentclass[letterpaper]{article} 
\usepackage{aaai25}  
\usepackage{times}  
\usepackage{helvet}  
\usepackage{courier}  
\usepackage[hyphens]{url}  
\usepackage{graphicx} 
\usepackage{natbib}  
\usepackage{caption} 
\usepackage{algorithm}
\usepackage{algorithmic}

\usepackage{booktabs}
\usepackage{amssymb} 
\usepackage{newfloat}
\usepackage{listings}
\DeclareCaptionStyle{ruled}{labelfont=normalfont,labelsep=colon,strut=off} 
\floatstyle{ruled}
\newfloat{listing}{tb}{lst}{}
\floatname{listing}{Listing}
\title{FastLGS: Speeding up Language Embedded Gaussians with Feature Grid Mapping}
\author{
    Yuzhou Ji\textsuperscript{\rm 1}\equalcontrib,
    He Zhu\textsuperscript{\rm 1}\equalcontrib,
    Junshu Tang\textsuperscript{\rm 2},
    Wuyi Liu\textsuperscript{\rm 1},\\
    Zhizhong Zhang\textsuperscript{\rm 1,3},
    Xin Tan\textsuperscript{\rm 1}\thanks{Corresponding Author.},
    Yuan Xie\textsuperscript{\rm 1}
}
\affiliations{
    \textsuperscript{\rm 1}East China Normal University, 
    Shanghai, China\\
    \textsuperscript{\rm 2}Shanghai Jiao Tong University, 
    Shanghai, China\\
    \textsuperscript{\rm 3}Shanghai Key Laboratory of Computer Software Evaluating and Testing

  xtan@cs.ecnu.edu.cn\\
  \textbf{https://george-attano.github.io/FastLGS}
    
}

\begin{document}

\maketitle

\begin{abstract}
The semantically interactive radiance field has always been an appealing task for its potential to facilitate user-friendly and automated real-world 3D scene understanding applications. 
However, it is a challenging task to achieve high quality, efficiency and zero-shot ability at the 
same time with semantics in radiance fields. In this work, we present FastLGS, 
an approach that supports real-time open-vocabulary query within 3D Gaussian Splatting (3DGS) under high resolution. 
We propose the semantic feature grid to save multi-view CLIP features which are extracted based on Segment Anything Model (SAM) masks, and map the grids to low dimensional features for semantic field training through 3DGS. Once trained, we can restore pixel-aligned CLIP embeddings through feature grids from rendered features for open-vocabulary queries. 
Comparisons with other state-of-the-art methods prove that FastLGS can achieve the first place performance concerning both \textbf{speed} and \textbf{accuracy}, where FastLGS is 98 $\times$ faster than LERF, 4 $\times$ faster than LangSplat and 2.5 $\times$ faster than LEGaussians. 
Meanwhile, experiments show that FastLGS is adaptive and compatible with many downstream tasks, 
such as 3D segmentation and 3D object inpainting, which can be easily applied to other 3D manipulation systems. 
\end{abstract}

%

\section{Introduction}

With the potential in fields such as robotics and augmented reality, 3D scene understanding has always been a conspicuous topic in the research community. 
Given a set of posed images, the goal is to learn an effective and efficient 3D semantic representation along with scene reconstruction, which should naturally support a wide range of downstream tasks in scene manipulation. 
However, it remains a vital challenge 
to simultaneously achieve \textbf{accuracy}, \textbf{efficiency} and the grounding of \textbf{open-vocabulary} level semantics together. 

Predominantly, existing methods usually only focus on the accuracy of differentiating objects, which 
normally assign points with labels or low-level features. {For example,}
Semantic-NeRF \cite{Zhi_2021_ICCV}, 
N3F \cite{10044452} and Panoptic Lifting \cite{Siddiqui_2023_CVPR} successfully 
lift labels or noisy 2D features into 3D radiance fields, 
providing semantic segmentations in certain scenes. 
However, their labels and low-level features {(e.g., DINO \cite{caron2021emerging} features)} are not sufficient for natural modality interactions {since some methods require extra position information (e.g., N3F \cite{10044452}) and others even do not provide any interaction (e.g., Semantic-NeRF \cite{Zhi_2021_ICCV} and Panoptic Lifting \cite{Siddiqui_2023_CVPR})}, 
and they also suffer from poor quality concerning in-the-wild scenes {due to the lack of zero-shot ability}. 
One possible solution for introducing zero-shot semantics {with friendly interactions} is to use {pre-trained text-image models with} high-level features. 
LERF \cite{Kerr_2023_ICCV}, as one of the state-of-the-art examples, enables pixel-aligned queries of the distilled 3D CLIP \cite{radford2021learning} embeddings, 
supporting long-tail open-vocabulary queries. 
Nevertheless, despite solving the problem of open-vocabulary interactions, LERF's results are  
completely not object-centric and can only extract fuzzy relevancy maps. 
The following LEGaussians \cite{shi2023language} proposes a dedicated embedding procedure to achieve smoother queries, but the results are still unstable and inconsistent. 
The above discussion shows the dilemma of balancing open-vocabulary ability and 
localizing accuracy when building a semantic field.

\begin{figure}[t]
   \centering
   \includegraphics[width = 0.8\columnwidth]{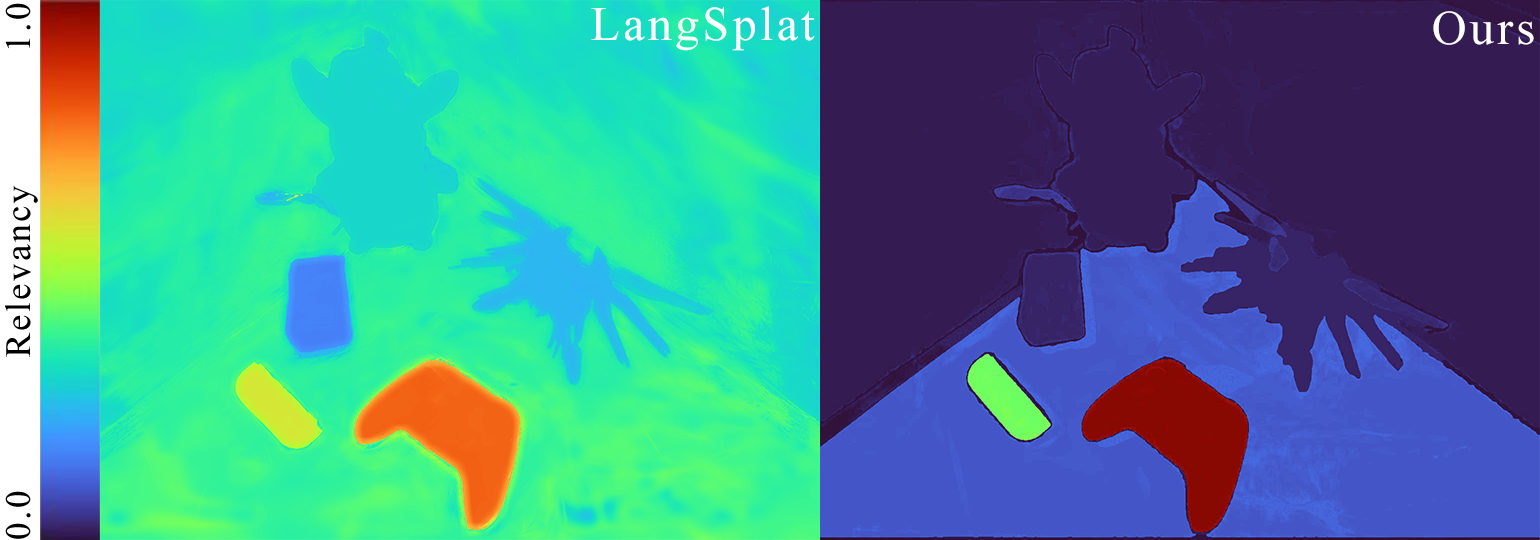}
   \caption{Visualized relevancy (turbo) of query ``Xbox wireless controller''. Our result is much more accurate in having higher relevancy within queried areas and lower relevancy among other regions compared with LangSplat.}
   \label{fig:intro}
\end{figure}

To get out of such a predicament, one may naturally think of achieving object-centric along with CLIP embeddings, but this brings numerous problems. 
First, we need to achieve the extraction of objects across all views to ensure object level 3D-consistency, where the popular Segment Anything Model (SAM) \cite{Kirillov_2023_ICCV} is strong enough to serve. 
However, to be a zero-shot method without any other model guidance in the whole 
process, the raw multi-granularity SAM masks can be confusing for the many-to-many 
masks mapping of even a single object. Recent LangSplat \cite{qin2023langsplat} trains an auto-encoder to handle this issue, but the reconstructed features are degraded and not sufficient for demanding query tasks. 
Meanwhile, directly building a high-level CLIP 
field is not only time-consuming but also inconsistent across views because different 
views of one object can provide close but distinct CLIP features. Moreover, the query 
strategy is also largely determined by the building of a semantic field. If we query by 
computing the CLIP relevancy of each pixel as in LERF, it is still not affordable. These are all key challenges for building the interactive 3D semantic field. 

Accordingly, in this paper, we present the FastLGS, which speeds up to build accurate 3D 
open-vocabulary semantic fields within 3D Gaussian Splatting (3DGS) \cite{kerbl3Dgaussians} via the feature grid mapping strategy. 
Given a set of posed images, we ensure object-centric by using a mask-based method as the 
basis. FastLGS first extracts all SAM masks and sent through the CLIP encoder to extract 
object-level image features. 
While directly training a CLIP feature field in high dimensionality could be time-consuming, we build a feature grid to map CLIP 
features to lower 3D space, but which inherently creates a many-to-one mapping problem. 
To tackle the problem of many-to-one mapping problem in 3D-consistent mapping, we propose a cross-view feature 
matching strategy to assign feature grids that represent scene objects with all sides 
of semantics. 
The low-dimensional mappings are attached to Gaussians to build a feature field. 
During the inference stage, pixel-aligned features will be rendered and we restore their 
semantics within feature grids to generate results upon open-vocabulary queries. 

During experiments, we found FastLGS can generate competitive target masks compared with state-of-the-art 3D segmentation and semantic field methods, grounding much more accurate language embeddings than 
LangSplat (see Figure 2). Meanwhile, the FastLGS is fast and supports 
real-time interactions under high resolution (98 $\times$ faster than LERF, 4 $\times$ faster than LangSplat and 2.5 $\times$ faster than LEGaussians, see Table \ref{tab:performance}), which can also be queried for single or multiple objects 
with similar queries in an adjustable way. Notably, FastLGS is adaptive and compatible with many downstream tasks that can be easily 
applied to systems like 3D segmentation and manipulation. 

In summary, the principal contributions of this work include:
\begin{itemize}
\item To our best knowledge, this is the first work that can interactively generate scene target masks upon natural language queries under high resolution (1440$\times$1080) in real-time (within one second). 
\item We propose the grid mapping strategy with the zero-shot open-vocabulary prompted paradigm for building a 3D semantic field in 3DGS. As far as we know, our method is one of the \textbf{fastest} and \textbf{most accurate} methods in generating pixel-aligned semantic features in the 3D space. 
\item We provide an efficient 3D semantic basis and find it adaptive to easily integrate with several downstream tasks like 3D segmentation and 3D object inpainting. 
\end{itemize}

\section{Related Work}
\subsection{NeRF and 3D Gaussian Splatting (3DGS)}
Neural Radiance Fields (NeRF) and 3D Gaussian Splatting (3DGS) have revolutionized 3D scene modeling and rendering. After first introduction, NeRF \cite{mildenhall2021nerf} has advanced largely with further innovations from mip-nerf \cite{Barron_2021_ICCV} to NeRF-MAE \cite{irshad2024nerfmae}. 3D Gaussian Splatting, detailed by \cite{kerbl3Dgaussians} and expanded upon in work \cite{luiten2023dynamic} on dynamic 3D Gaussians, optimizes the rendering of point clouds with Gaussian kernels for real-time applications. Later contributions by \cite{liu2024citygaussian} and others \cite{zhu2023FSGS} \cite{chen2024mvsplat} \cite{Tian2025drive} underscore the ongoing enhancements and versatility of 3DGS in handling increasingly complex rendering tasks. Noting the advantages of 3D Gaussian Splatting in rendering speed and quality, our paper chooses it as a representation for building semantic field.

\subsection{2D and 3D Segmentation}

The field of 2D image segmentation has undergone remarkable progress by the adoption of Transformer architecture, notably through SEgmentation TRansformer (SETR) \cite{Zheng_2021_CVPR}, alongside studies \cite{NEURIPS2021_950a4152,NEURIPS2021_64f1f27b,Cheng_2022_CVPR,sun2024uni}. 
While works like CGRSeg \cite{ni2024context} provides excellent accuracy and efficiency, 
innovations such as SAM \cite{Kirillov_2023_ICCV} and SEEM \cite{NEURIPS2023_3ef61f7e} utilize various kinds of prompts for segmentation, inspired many mask based researches. Meanwhile, image captioning has also become a promising method of retrieving image semantics \cite{NEURIPS2023_804b5e30}.

Advancements in 3D segmentation have paralleled those in 2D, 
with all kinds of innovations include Cylinder3D \cite{zhou2020cylinder3d} for LiDAR semantic segmentation in driving scenes, and 3D semantic segmentation within point clouds \cite{tan2023positive,sun2024image}. Other developments in radiance fields have also refined the precision and applicability of 3D segmentation techniques \cite{Goel_2023_CVPR,tang2023scene,cen2023segment}, but open-vocabulary text guided 3D segmentation remains a challenge for the lack of compatible semantic scene construction. 

\subsection{Semantic 3D scene}
Advancements in semantic integration have significantly enhanced 3D modeling, notably through NeRF and 3DGS. \cite{Zhi_2021_ICCV} introduced semantic layers into NeRF, setting the stage for enriched scene understanding. This has been further developed in studies such as LERF \cite{Kerr_2023_ICCV}, ISRF \cite{Goel_2023_CVPR}, DFF \cite{9812291}, and N3F \cite{10044452}, which integrate detailed semantic segmentation to refine 3D visualizations.  In special scenes, works such as MA-52 \cite{guo2024benchmarking} and COTR \cite{ma2024cotr} have also served as good examples.

In parallel, enhancements in 3DGS have also advanced semantic capabilities. The foundational work  \cite{kerbl3Dgaussians} and further developments \cite{gu2024egolifter,ye2023gaussian} have refined the application of 3DGS in semantic segmentation tasks. The SAGA framework \cite{cen2023saga} demonstrates the integration of detailed 2D segmentation outcomes into 3D models, enhancing semantic accuracy. LEGaussians \cite{shi2023language} and LangSplat \cite{qin2023langsplat} explore incorporating language processing into 3DGS, adding a new layer to semantic interpretations in 3D scene modeling. 
However, with inconsistent or fuzzy results in varying views and slow query speed, the quality of semantic fields built by these approaches still limits their further applications.

\begin{figure*}[!t]\centering
  \includegraphics[width=\textwidth]{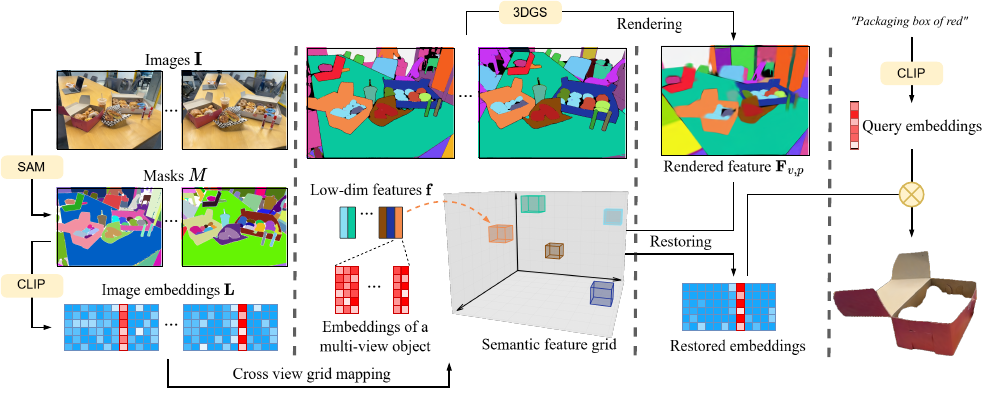}
  \caption{FastLGS pipeline. Left: Initialization. Mid: Feature grid construction and embeddings restoring. Right: query using open-vocabulary prompts.}
  \label{fig:pipeline}
\end{figure*}

\section{Method}
\subsection{Overview}
As in Figure \ref{fig:pipeline}, the input images are initially put for extracting SAM masks and CLIP embeddings. This information 
is utilized together for the building of a semantic feature grid and low-dim feature mapping.
Then, the obtained pixel-aligned low-dim features are learned in parallel with scene reconstruction according to ``Training Features for Gaussians'' where 
we describe the training for embedded features through Gaussians. 
At the inference stage, the built semantic field can be interactively queried for high quality masks in novel views using natural language based on restored embeddings. 
The query strategy is demonstrated in ``Querying Features''.

\subsection{Initialization}
\label{sec:INIT}
First, we initialize the original semantics. 
Given input images $\{\textbf{I}_t|t=0,1,...,T\}$, for each image $\textbf{I}_i$, we utilize SAM to obtain $whole$ segmentation masks $\{\textit{M}_{i,j}|j=0,1,...m_i\}$ with a regular grid of 32 $\times$ 32 point prompts and compute respective CLIP features $\{\textbf{L}_{i,j}|j=0,1,...m_i\}$. 
Masks and CLIP embeddings will be used for building feature grids along with 
pixel-aligned low-dim feature maps. After initialization, no other pre-trained model will be required 
except for a CLIP text encoder at the inference stage. 

\subsection{Semantic Feature Grid}
\label{sec:feature grid}
By practice, simply integrating CLIP feature with 3DGS can lead to infeasible demand of memory and limited efficiency of rendering as each Gaussian needs to record and update a parameter in hundreds of dimensions (See ablations). 
On the other hand, the CLIP features of an object could show unexpected variance due to the view angles, redundant background and complex occlusion, posing a challenge to maintaining consistency and accuracy of semantics in 3D scenes. 
As language features of the same object from multiple views should share similar semantics, it could be compressed for lower training and query costs. However, current MLP-based compressed and reconstructed features cannot provide queries of higher quality as the former can be inconsistent and inaccurate like the raw features while the latter can be degraded which is not sufficient for demanding scenes (see relevancy maps in Figure \ref{fig:intro}). Meanwhile, similar methods also can not acquire unseen features in a single view, restricting the possibility of querying objects with semantics captured in 
another view but invisible through the current angle. 

To address these issues, we propose semantic feature grid, which stores the multi-view language features of each object in the scene and is mapped to a low dimensional feature. Specifically, our goal is to map $\textbf{L}\in\mathbb{R}^D$ to $\textbf{f}\in(0,1)^d$ where language features $\textbf{L}$ of the same object share the same low-dim feature $\textbf{f}$. Each feature $\textbf{f}$ is assigned to a grid with the size of $\mathcal{K}^{-\frac{1}{d}}$, where $\mathcal{K}$ refers to the number of objects in a specific scene computed during cross view matching. 
The low dimensional features will be assigned pixel-wise according to corresponding masks and trained efficiently in 3DGS, while the grid can be used to restore accurate CLIP embeddings based on rendered low-dim features during inference. In practice, this strategy enables our method to achieve less consumption and faster speed (Table \ref{tab:performance}). 

\subsection{Cross View Grid Mapping}
\label{sec:matching}
To achieve consistent features of objects across views, we sequentially match the denoised masks in adjacent image set $\textbf{I}$ and assign features. Considering the possible instability of CLIP features in multi-views mentioned previously, the matching process consists of both key points correspondence and feature similarity comparison.

\textbf{Key points correspondence} 
We prioritize the mask correspondence by key points obtained using the SIFT and the k-nearest neighbors (KNN) algorithm for accurate and robust matching. The SAM masks with more than $\tau$ corresponding key point pairs will be assigned to the same low dimensional feature $\textbf{f}$ and considered as segmentation masks of an object from different viewpoints.

\textbf{Feature similarity comparison}
As some segmentations with smooth pixel value have few key points, the matching process is employed according to similarity calculated by hybrid features of CLIP embedding $\textbf{L}$ and color distribution $\textbf{C}$. For $\textbf{I}_i$ and $\textbf{I}_j$ with $m_i$ and $m_j$ mismatched masks, we calculate the similarity matrix $\textbf{SIM}_{m_i\times m_j}$ and choose the mask pairs with highest similarity. A threshold $\theta$ is set that masks with similarity $sim$ higher than $\theta$ will have the same feature $\textbf{f}$ and the others will be assigned to a new low dimensional feature.

For two segmentation masks $\textit{M}_i$, $\textit{M}_j$, the similarity $sim_{i, j}$is calculated as follow:
\begin{equation}
\mathbf{sim}_{i,j}=\alpha{sim}_{i, j}^{color}+(1-\alpha){sim}_{i,j}^{CLIP},
\end{equation}
where $\alpha$ is the weight of similarity of color distribution features $\textbf{C}$. 
${sim}_{i, j}^{CLIP}$ is calculated based on Bhattacharyya distance of $\textbf{C}$ while ${sim}_{i, j}^{color}$ is calculated by cosine similarity:
\begin{equation}
sim_{i, j}^{CLIP}=\frac{\textbf{L}_{i} \cdot \textbf{L}_{j}}{\left\|\textbf{L}_{i} \right\| \left\|\textbf{L}_{j} \right\|}
;\hspace{2mm}
sim_{i, j}^{color}=\sqrt{\textbf{C}_i \textbf{C}_j}.
\end{equation}

Algorithm \ref{alg:one} displays our procedure of cross view grid mapping, where we use a variable $Idx$ to refer to low-dim feature index of each mask. The consistency of objects across views is built and the amount $\mathcal{K}$ is accessed during the matching process, which helps establish our semantic feature grids and low-dimensional feature mapping. 

\begin{algorithm}[tb]
\caption{Cross View Grid Mapping}
\label{alg:one}
\textbf{Input}: Image sequence $\{\textbf{I}_t|t=0,1,...,T\}$, segmentation masks $\{\textit{M}_{i,j}|i=0,1,...,T;\:j=0,1,...,m_i\}$, CLIP embedding $\{\textbf{L}_{i,j}|i=0,1,...,T;\:j=0,1,...,m_i\}$ and color distribution $\{\textbf{C}_{i,j}|i=0,1,...,T;\:j=0,1,...,m_i\}$, index of mask $M_{i, j}$'s low dimensional feature $Idx_{i,j}$, threshold $\tau$ and $\theta$\\
\textbf{Parameter}: index of image $i$, mask's index of image $\textbf{I}_i$'s mask $j$, index of the most corresponding mask $k$, images for matching $\textbf{I}'$, masks for matching $\textit{M}'$\\
\textbf{Output}: low dimensional feature indices $Idx$ of masks $\textit{M}$ and the number $\mathcal{K}$ of objects in a scene
\begin{algorithmic}[1] 
\STATE $\mathcal{K}$, $Idx_{0}$ = initialise($\textbf{I}_0$, $\textit{M}_0$, $\textbf{L}_0$)
\STATE $\textbf{I}'$ = $\{\textbf{I}_0\}; \textit{M}'$ = $\{\textit{M}_{0,i}|i=0,1,...,m_0\}$.
\STATE Let $i=1$.
\WHILE{$i \leq T$}
\STATE $corrInfo$ = correspondKp($I_i$, $\textbf{I}'$, $\textit{M}'$, )
\WHILE{$j \leq m_i$}
\STATE $k$, $\textit{M}'_k$, $numKp$ = selMostCorrMask($\textit{M}_{i,j}$, $corrInfo$)

\IF{$numKp \ge \tau$}
\STATE $Idx_{i,j}$ = low-dim feature index of $\textit{M}'_k$\;
\ELSE
\STATE $sims_{i,j}$ = computeSims($\textit{M}_{i,j}$, $\textit{M}'$, $\textbf{L}$, $\textbf{C}$)
\STATE $k$, $\textit{M}'_k$, $sim$ = selMostSimMask($\textit{M}_{i,j}$, $\textit{M}'$, $sims_{i,j}$)

\IF{$sim \ge \theta$}
\STATE $Idx_{i,j}$ = low-dim feature index of $\textit{M}'_k$\;
\ELSE 
\STATE $\mathcal{K}$ = $\mathcal{K} + 1$; $Idx_{i,j}$ = $\mathcal{K}$
\ENDIF
\ENDIF
\STATE $\textbf{I}'$ = union($\textbf{I}'$, $\{\textbf{I}_i\}$)\;
\STATE $\textit{M}'$ = union($\textit{M}'$, $\{\textit{M}_{i,j}|j=0,1,...,m_i\}$)\;
\STATE $j = j+1$
\ENDWHILE
\STATE $i=i+1$
\ENDWHILE
\end{algorithmic}
\end{algorithm}

\subsection{Training Features for Gaussians}
\label{sec:train&loss}

We train a mapped low-dimensional feature $\mathbf{f}_{m}$ for each gaussian $g$ to build a feature field, 
which is adaptive for integration. 

\textbf{Rendering Features} Given a camera pose $v$, we compute the feature $\mathbf{F}_{v,p}$ of a pixel 
by blending a set of ordered Gaussians $\mathcal{N}$ overlapping the pixel similar to the color computation 
of 3DGS: 
\begin{equation}
\mathbf{F}_{v,p}=\sum_{i\in\mathcal{N}}\mathbf{f}_{i}a_i\prod\limits_{j=1}^{i-1}(1-a_j),
\end{equation}
where $a_i$ is given by
evaluating a 2D Gaussian with covariance $\sum$ multiplied with
a learned per-Gaussian opacity.

\textbf{Optimization} The low-dim features follows 3DGS optimization pipeline and especially inherits the 
fast rasterization for efficient optimization and rendering. The loss function for features 
is $\mathcal{L}_1$ combined with a D-SSIM term:
\begin{equation}
    \mathcal{L}_f=(1-\lambda)\mathcal{L}_1+\lambda\mathcal{L}_{D-SSIM},
\end{equation}
where $\lambda$ is also fixed to 0.2 in all cases. 

\subsection{Querying Features}
\label{sec:query}
Once trained, FastLGS can be interactively queried for different objects in the scene using open vocabulary 
text prompts. 
First, we query the pixel-aligned semantics $\mathbf{F}_v$ by projecting the attached feature to view the plane using the 
rendering for features mentioned earlier, which can be used for all queries 
within camera pose $v$. 

\textbf{Relevancy Score} When provided a prompt, we assign its language relevancy scores 
based on grid image features similar to LERF. 
We compute the cosine similarity between image embedding $\phi_{img}$ and canonical phrase embeddings 
$\phi_{canon}^i$, then compute the pairwise softmax between image embedding and text prompt embedding 
$\phi_{query}$, so that the relevancy score is: 
\begin{equation}
\label{eql:relev}
S_{relev}=\min_i\frac{\exp(\phi_{img}\cdot\phi_{query})}{\exp(\phi_{img}\cdot\phi_{canon}^i)+\exp(\phi_{img}\cdot\phi_{query})} .
\end{equation}

For canonical phrases, we use ``object'', ``stuff'' and ``texture'' for all queries.

\textbf{Target Mask} We go through feature grids to restore high-level 
image features and use them to compute relevancy with text embeddings. The grid with the 
highest relevancy score is deemed to be the target grid queried. 
Because each grid has multi-view image features of an object, we are even able to locate targets based on 
semantics invisible in the current query view (for example, locating a book named ``\textit{computer vision}'' 
from the back of it). 
While the rendered semantics have been smoothed, we locate their corresponding 
feature grids by calculating Euclidean distance between grids and rendered features: 
$Dis_{gf}=\sqrt{\sum_{i=0}^{2} (\mathbf{F}^{v,i}_p-\mathbf{p}_{i})^2}$
where $\mathbf{p}$ is the grid's corresponding low-dim features. 
We compute pixel-wise $Dis_{gf}$ based on target grid, and points with $Dis_{gf}$ lower than 
threshold $\tau_{ac}$ forms the target mask. 
By changing the number of grids, we can adjustably query multiple targets. 

\section{Experiments}
\label{sec:exp}

In this section, we first show the speed and quality of 
open-vocabulary object retrieval in 
comparison with other state-of-the-art methods through quantitative experiments, 
then we provide qualitative results of downstream tasks including 3D segmentation and object deletion. 
Ablation studies are conducted to demonstrate the rationality of feature grid based design. 

\textbf{Datasets} For quantitative experiments, we train and evaluate the models on datasets including 
SPIn-NeRF \cite{mirzaei2023spin}, LERF \cite{Kerr_2023_ICCV} and 3D-OVS \cite{liu2023weakly}.

\textbf{Implementation Details} We use the same OpenClip ViT-B/16 model as LERF and SAM ViT-H model 
as LangSplat. We train the features and scenes in 3DGS for 30,000 iterations. 
Activation threshold $\tau_{ac}$, correspondence threshold $\tau$ and $\theta$ are set to $5.0$, 4 and 0.95. Weight $\alpha$ is set to 0.3. $\textbf{f}$ is normalized to $(0,255)^3$. 
Experiments show robustness of the proposed method to the aforementioned parameters. 
While the original time calculation of LangSplat puts aside feature rendering time and feature reconstructing time, here we compute the query time of the whole query process as LERF does. 
The tested LERF masks are regions with relevancy higher than 20\% after normalization. 
The normalization for each query is from 50\% (less relevant than canonical phrases) 
to the maximum relevancy, which is identical to the visualization strategy of LERF. 
All results are reported running on a single TITAN RTX GPU.

\begin{figure*}[!t]\centering
  \includegraphics[width=\textwidth, height=15cm]{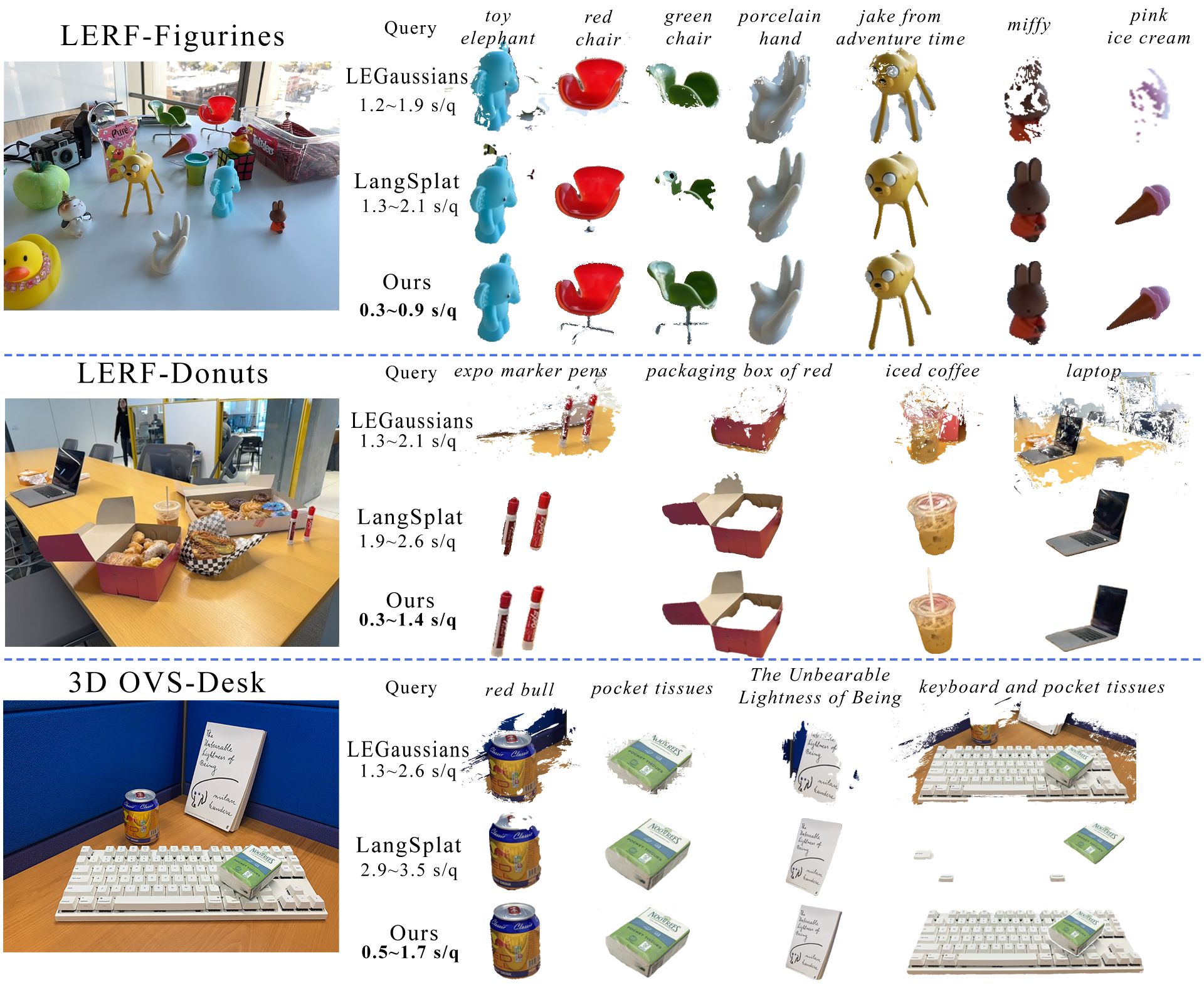}
  \caption{Visual results of retrieved objects in different scenes.}
  \label{fig:qualitative}
\end{figure*}

\subsection{Comparison With Other State-of-the-art Methods}
\label{sec:Qresult}

For quantitative experiments, we test different methods 
on the SPIn-NeRF dataset, the LERF dataset and 3D-OVS dataset. 
We show the quality of FastLGS-generated masks by comparing with other 
state-of-the-art 3D segmentation methods including the multi-view segmentation of SPIn-NeRF (MVSeg) \cite{mirzaei2023spin} and SA3D \cite{cen2023segment}, 
and compare the language retrieval ability with LERF \cite{Kerr_2023_ICCV}, 
LEGaussians \cite{shi2023language} and LangSplat \cite{qin2023langsplat}. 

\textbf{SPIn-NeRF dataset} 
We evaluate the IoU and pixel-wise accuracy of masks with provided ground truth (1008 $\times$ 567), and also show the time consumption for each query, which is omitted in MVSeg and SA3D because they do not support single view query. 
For LERF, LangSplat, LEGaussians and FastLGS, we use the same text queries for the target segmentation objects to generate masks. 
Other methods all follow their original settings when tested on this dataset. 
We also provide the 2D segmentation results generated by SAM based on manual point prompts of 
original scene images for comparison. 
As shown in Table \ref{tab:mvseg}, FastLGS generates masks with competitive quality which have higher mIoU compared with other methods, 
and largely outperforms relevant methods in query speed.

\textbf{LERF dataset} The LERF dataset contains a number of in-the-wild scenes and is 
much more challenging, which strongly requires zero-shot abilities.  
We report localization accuracy for the 3D object localization task following LERF \cite{Kerr_2023_ICCV} with 
ground truth annotations provided by LangSplat \cite{qin2023langsplat} (resolution around 985 $\times$ 725). 
Results are shown in Table \ref{tab:lerf} and visual examples in Figure \ref{fig:qualitative} (line 1\&2), which further demonstrates 
FastLGS's advantages in natural language retrieval. 

\textbf{3D-OVS dataset} We also compare with 2D-based open-vocabulary segmentation 
methods including ODISE \cite{xu2023open} and OV-Seg \cite{liang2023open} along with 3D-based methods including 3D-OVS \cite{liu2023weakly}, LERF \cite{Kerr_2023_ICCV}, LEGaussians \cite{shi2023language} and 
LangSplat \cite{qin2023langsplat}. Results are provided in Table \ref{tab:ovs}, where our method can outperform both 2D and 3D methods. 
In Figure \ref{fig:qualitative} (line 3) our method can also better support demanding custom query 
``keyboard and pocket tissues'' which fails in LangSplat. 
Table \ref{tab:performance} further shows the query time 
and model size where our method also outperforms other methods, proving its advantages in real-world applications.

\begin{table}
\centering
\begin{tabular}{cccc}
  \toprule
  Method          & mIoU (\%)  &mAcc (\%) &mTime (s)\\
  \midrule
  SAM(2D)  & 95.7 & 96.1 &0.05\\
  \midrule
  MVSeg     & 89.5 & 97.8 &-\\
  SA3D  & 90.9 & 98.3 &-\\
  LERF  & 81.0 & \textbf{99.5} &30.2\\
  LEGaussians & 89.3 & 97.8 & 1.04 \\
  LangSplat & 92.2 & 94.7 &1.43\\
  OURS & \textbf{93.1} & 95.2 &\textbf{0.31}\\
  \bottomrule
\end{tabular}
\caption{Quantitative Results on SPIn-NeRF dataset (LERF masks usually cover large surrounding areas and result in high mAcc and low mIoU).}
\label{tab:mvseg}
\end{table}

\begin{table}%
\centering
\begin{tabular}{ccccc}
  \toprule
  Scene    & LERF & LEG & LangSplat & OURS\\
  \midrule
  ramen          & 61.9 & 78.6 & 73.2 & \textbf{84.2}\\
  figurines      & 75.5 & 73.7 & 80.4 & \textbf{91.4}\\
  teatime        & 84.8 & 85.6 & 88.1 & \textbf{95.0}\\
  waldo\_kitchen & 70.2 & 90.1 & 95.5 & \textbf{96.2}\\
  \midrule
  Overall        & 73.1 & 82.0 & 84.3 & \textbf{91.7}\\
  \bottomrule
\end{tabular}
\caption{Quantitative Results of localization accuracy on LERF dataset. (LEGaussians as LEG)}
\label{tab:lerf}
\end{table}%

\begin{table}%
\centering
\begin{tabular}{ccccccc}
  \toprule
  Method    & \textit{bed} & \textit{bench} & \textit{room} & \textit{sofa} & \textit{lawn} & Overall\\
  \midrule
  ODISE     & 55.6 & 30.1 & 53.5 & 49.3 & 39.1 & 45.5\\
  OV-Seg    & 79.8 & 88.9 & 71.4 & 66.1 & 81.2 & 77.5\\
  \midrule
  3D-OVS    & 89.5 & 89.3 & 92.8 & 74.1 & 88.2 & 86.8\\
  LERF      & 76.2 & 59.1 & 56.4 & 37.6 & 78.2 & 61.5\\
  LEGaussians & 45.7 & 47.4 & 44.7 & 48.2 & 49.7 & 47.1\\  
  LangSplat & 92.6 & 93.2 & 94.1 & 89.3 & 94.5 & 92.7\\
  \midrule
  OURS      & \textbf{94.7} & \textbf{95.1} & \textbf{95.3} & \textbf{90.6} & \textbf{96.2} & \textbf{94.4}\\
  \bottomrule
\end{tabular}
\caption{Quantitative Results of mIoU scores (\%) on 3D-OVS dataset.}
\label{tab:ovs}
\end{table}%

\begin{table}%
\centering
\begin{tabular}{ccccc}
  \toprule
  Scene ``sofa''    & LERF & LangSplat & LEG & OURS\\
  \midrule
  mTime (s/q)     & 51.2 & 2.14 & 1.31 & \textbf{0.52}\\
  Model Size      & 1.28GB & 565MB & 585MB & \textbf{200MB}\\
  \bottomrule
\end{tabular}
\caption{Comparison of performance and consumption (time computed on rendering resolution 1440x1080).}
\label{tab:performance}
\end{table}%

\subsection{Downstream Applications}
\label{sec:qualitative}
In this section, we test our method for the ability of applying to downstream 3D object manipulation tasks. 

\textbf{Language Driven 3D Segmentation} We integrate FastLGS with Segment Any 3D Gaussians (SAGA) 
\cite{cen2023saga} to achieve language driven 3D segmentation. 
The original SAGA requires manual positional prompting for reference mask selection, 
here we can directly use our masks as segmentation references and 
conduct 3D segmentation based on natural language.
The whole query and segmentation process can be completed around only one second. 

\textbf{Language Driven Object Inpainting} We generate target masks of queried objects across views and use them as multi-view segmentation masks. 
We follow the multi-view segmentation-based 
inpainting pipeline of SPIn-NeRF \cite{mirzaei2023spin} to conduct 3D object inpainting using the extracted 
multi-view segmentation masks. The masks are dilated by default with a 5x5 kernel for 5 iterations to ensure that all objects are masked as in SPIn-NeRF. 
Results show that consistent masks can be directly used for high-quality object inpainting 
through 3D inpainting methods. 

According to the above experiments, our method proves to be superior in grounding 
accurate semantics with interactive efficiency and more affordable consumption, further 
paving the way for many downstream 3D manipulation tasks in open-vocabulary interaction works. 

\subsection{Ablation studies}
\label{sec:ablation}
We conduct ablation to validate the necessity of our semantic feature grid for query and the improvements for the accuracy of our components in building cross-view grid mapping.

\textbf{Querying Feature}
The performance is reported in Table \ref{tab:ablation_query}, where SFG represents the semantic feature grid. Initially, we directly reconstruct the semantic field with NeRF using extracted CLIP features for SAM-based masks of images. When replacing NeRF with 3D Gaussian Splatting, the high dimensions of CLIP feature result in running out of memory. The issue is addressed by combining it with a semantic feature grid, which also brings improvements in accuracy and speed.

\begin{figure}[h]\centering
  \includegraphics[width=0.9\columnwidth]{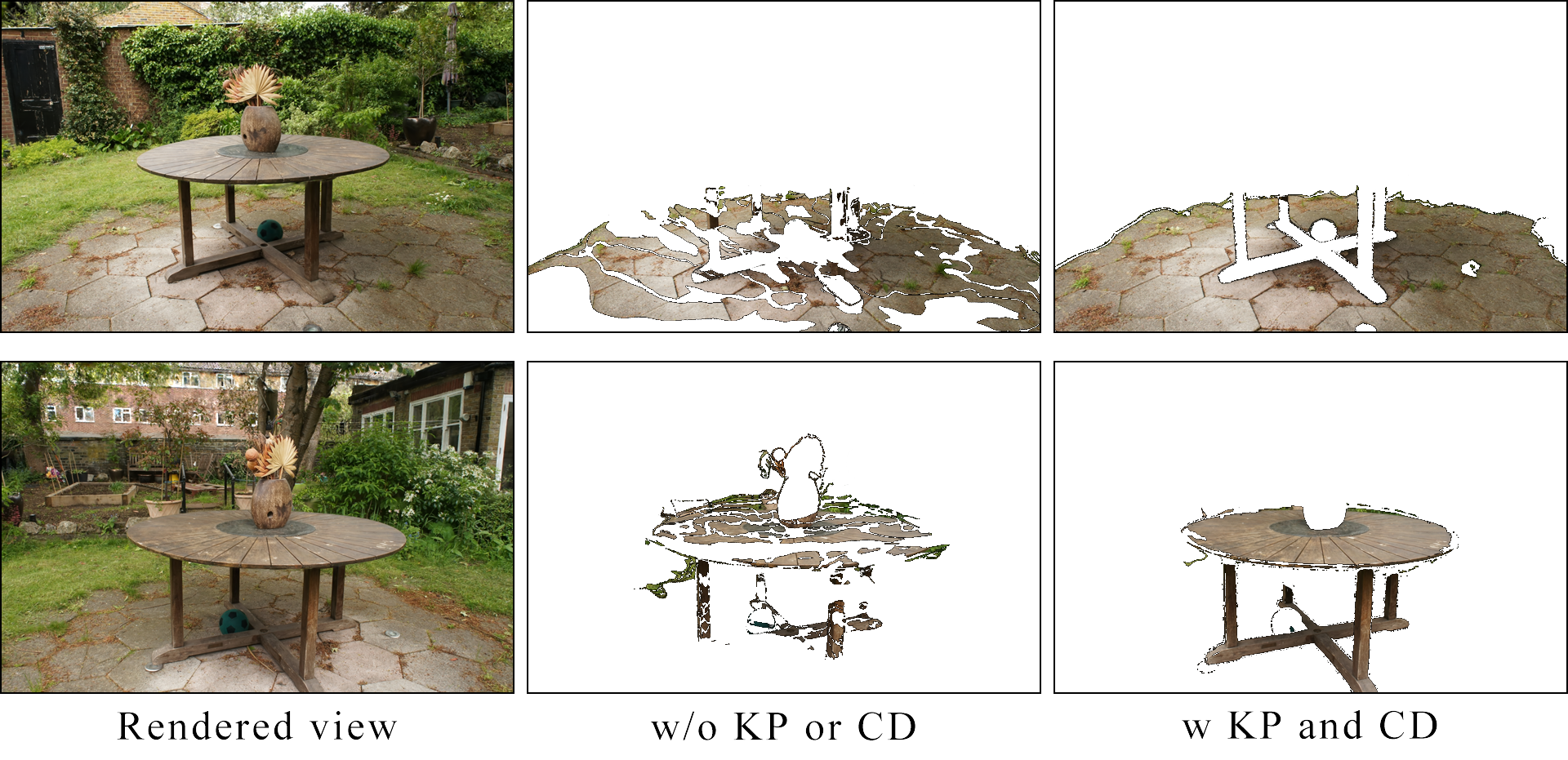}
  \caption{Ablation on keypoint and color feature matching.}
  \label{fig:ablation}
\end{figure}

\textbf{Cross View Matching}
We report results in Table \ref{tab:ablation_grid} and Figure \ref{fig:ablation}, where KP represents key points and CD represents color distribution feature \textbf{C}.

Key points correspondence and features similarity with the auxiliary of color distribution provide more guidance for maintaining consistency of objects across a series of continuous perspectives, which are dedicated to training low-dim features of high quality for Gaussians and effectively improving the performance. 

\begin{table}%
\centering
\begin{tabular}{ccccc}
  \toprule
  \multicolumn{3}{c}{Component} & \multicolumn{2}{c}{Performance}\\
  \midrule
  SAM & 3D-GS & SFG & mIoU(\%) & mTime(s) \\
  \checkmark & & & 83.3 & 20.1 \\
  \checkmark & \checkmark& & OOM & OOM \\
  \checkmark & \checkmark & \checkmark & 95.1 & 0.98 \\
  \bottomrule
\end{tabular}
\caption{Ablation on querying feature.}
\label{tab:ablation_query}
\end{table}%

\begin{table}%
\centering
\begin{tabular}{ccccc}
  \toprule
  \multicolumn{3}{c}{Component} & \multicolumn{2}{c}{Performance}\\
  \midrule
  CLIP & KP & CD & mIoU(\%) & mAcc(\%) \\
  \checkmark & & & 75.2 & 83.2 \\
  \checkmark & \checkmark& & 87.5 & 91.8 \\
  \checkmark & \checkmark & \checkmark & 92.1 & 97.5 \\
  \bottomrule
\end{tabular}
\caption{Ablation on the semantic feature grid.}
\label{tab:ablation_grid}
\end{table}%

\section{Conclusion}

In this paper, we propose FastLGS, a method that speeds up to build accurate 3D 
open-vocabulary semantic fields within 3DGS. By mapping and restoring multi-view 
CLIP embeddings through feature grids instead of MLPs, FastLGS not only has interactive 
query speed, but also provides more accurate target masks and semantic relevancy 
along with lower consumption compared with state-of-the-art methods. 
Meanwhile, experiments show FastLGS's ability to be 
integrated with other downstream 3D manipulation tasks. 
While FastLGS now follows an object-centric paradigm, challenges occur in querying for parts of an object, such as ``bicycle handlebars'' or ``chair leg''. We believe this could be solved by building the feature grid in a multi-granularity way, where actual performance along with cross-view 
consistency of both mask extraction and query results remains a problem, which requires further research.

\section{Acknowledgments}
This work was supported in part by the National Natural Science Foundation of China (62302167, U23A20343, 62222602, 62176092, and 62476090); in part by Shanghai Sailing Program (23YF1410500); Natural Science Foundation of Shanghai (23ZR1420400); in part by Chenguang Program of Shanghai Education Development Foundation and Shanghai Municipal Education Commission (23CGA34), in part by CCF-Tencent RAGR20240122.

\bibliography{aaai25}

\end{document}